\documentclass[sigconf,nonacm]{acmart}
\AtBeginDocument{%
  }

\usepackage{xspace,xpunctuate}
\usepackage{dialogue}

\newcommand{\demourl}{{http://alphaevolve-examples.web.app}}
\newcommand{\setupagent}{{experiment configuration assistant}}
\newcommand{\numplaytesters}{{11}}
\newcommand{\numexperiments}{{2,300}}

\newcommand{\playtestduration}{{three months}}
\newcommand{\playtestname}{{expert evaluation}}
\newcommand{\playtestername}{{participant}}
\newcommand{\numactives}{{150}}
\newcommand{\plainquote}[1]{{\emph{``#1''}}}
\newcommand{\ie}{{i.e.,}\xspace}
\newcommand{\cf}{{cf.}\xspace}
\newcommand{\eg}{{e.g.,}\xspace}
\newcommand{\ea}{{et~al\xperiod}\xspace}

\graphicspath{{figures/}} 

\begin{document}

\title[Intentmaking and Sensemaking]{Intentmaking and Sensemaking:\texorpdfstring{\\}{ }Tightening the Scientific Loop in AI-Guided Mathematical Discovery}

\author{Alex Bäuerle}
\authornote{Both authors contributed equally to this research.}
\orcid{1234-5678-9012}
\affiliation{%
  \institution{Google DeepMind}
  \city{Paris}
  \country{France}}

\author{Adam Connors}
\authornotemark[1]
\affiliation{%
  \institution{Google DeepMind}
  \city{London}
  \country{United Kingdom}}
  
\author{Alexander Novikov}
\affiliation{%
  \institution{Google DeepMind}
  \city{London}
  \country{United Kingdom}}
  
\author{Adam Zsolt Wagner}
\affiliation{%
  \institution{Google DeepMind}
  \city{London}
  \country{United Kingdom}}
  
\author{Ngân Vũ}
\affiliation{%
  \institution{Google DeepMind}
  \city{London}
  \country{United Kingdom}}
  
\author{Fernanda Vi\'egas}
\affiliation{%
  \institution{Google DeepMind}
  \city{Cambridge}
  \country{USA}}
  
\author{Martin Wattenberg}
\affiliation{%
  \institution{Google DeepMind}
  \city{Cambridge}
  \country{USA}}
  
\author{Lucas Dixon}
\affiliation{%
  \institution{Google DeepMind}
  \city{Paris}
  \country{France}}

\renewcommand{\shortauthors}{Bäuerle and Connors et al.}

\begin{abstract}
  Artificial intelligence offers powerful paradigms for scientific discovery, orchestrating autonomous inner search loops over long horizons.
  However, discovery remains bottlenecked by the friction of the outer human-AI loop: the cycle where domain experts formulate research objectives, diagnose reward hacks, and extract meaning.
  We present findings from a study with \numplaytesters{} expert mathematicians using AlphaEvolve, an evolutionary coding agent, across more than \numexperiments{} experiments over \playtestduration{}.
  We identify and characterize \emph{intentmaking}, the iterative process of discovering, defining, and refining experimental goals through system interaction, as a reciprocal counterpart to \emph{sensemaking}.
  We observe that domain experts effectively alternate between steering the search space (intentmaking) and extracting mathematical insights (sensemaking).
  Our findings suggest an approach to designing AI tools for scientific discovery that moves beyond the existing question--answer model, treating them as collaborative instruments rather than opaque black-box oracles.
\end{abstract}



\begin{teaserfigure}
  \includegraphics[width=\textwidth]{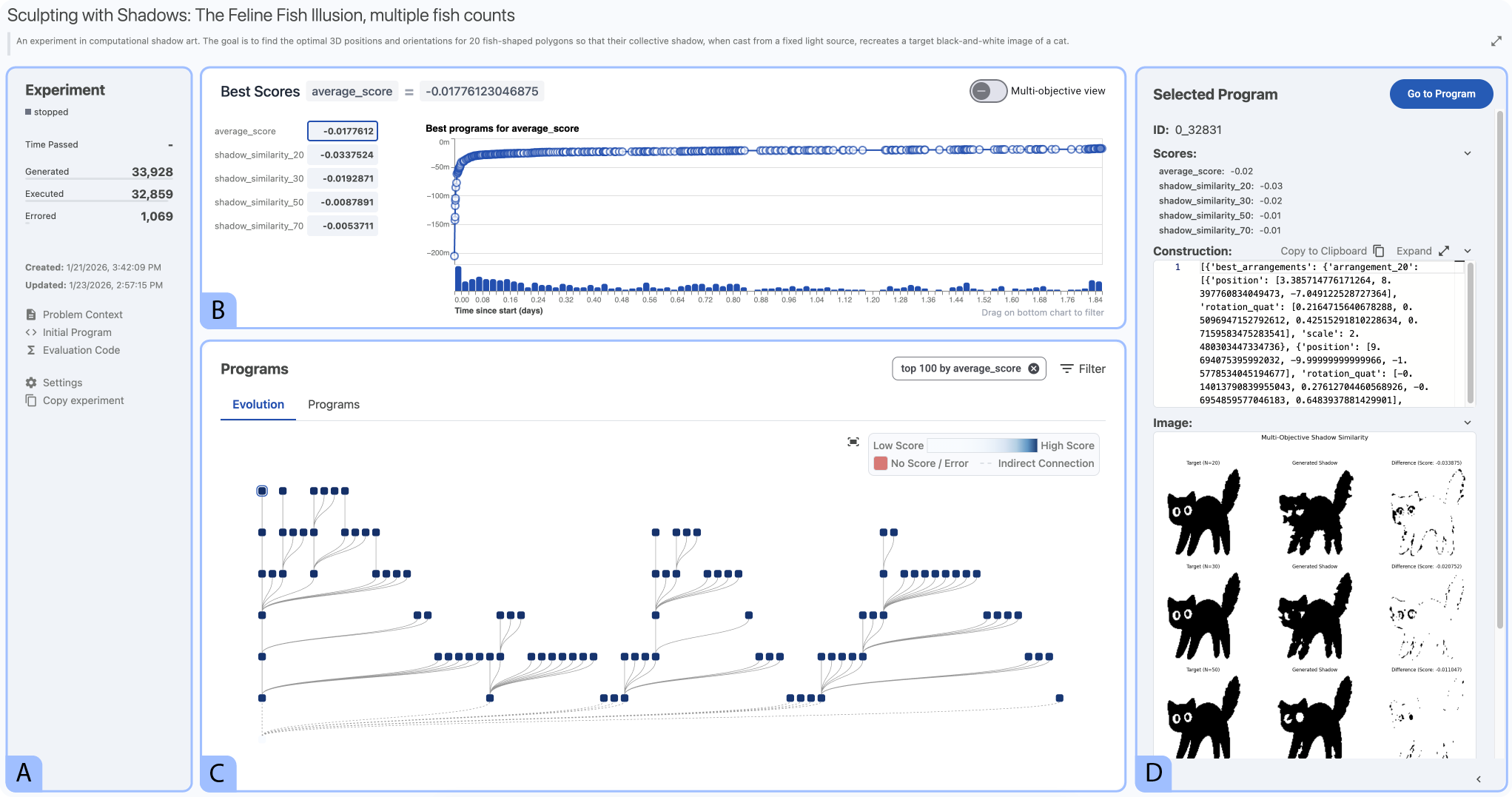}
  \caption{Sensemaking Dashboard for an AlphaEvolve experiment.
  (A) Overview of the experiment's progress, number of generated candidates, and authorship information.
  (B) Best scores, the current best metrics per score and the progress for the selected score.
  (C) Programs view, showing the evolution tree of candidates.
  (D) Selected program sidebar, a quick overview of the selected program's characteristics and artifacts.
  Interactive examples of AlphaEvolve experiments can be found at: \demourl{}.} 
  \Description{Dashboard for an AlphaEvolve experiment, showing four key panels: (A) experiment summary statistics and candidate generation counters; (B) best score progression charts over time; (C) an evolution tree visualization of candidate programs; and (D) a selected program sidebar showing program attributes, construction metrics, and visual artifacts.}
  \label{fig:dashboard}
\end{teaserfigure}

\maketitle

\section{Introduction}

The integration of artificial intelligence into scientific and mathematical discovery has generated immense excitement, promising to accelerate breakthroughs across numerous domains~\cite{wang2023scientific,romera2024mathematical}.
In mathematics in particular, recent AI-driven breakthroughs have catalyzed intense debate within the community~\cite{castelvecchi2023will,tao2026mathematics}.
While modern automated systems demonstrate remarkable aptitude at discovering counterexamples, optimizing combinatorial bounds, and completing formal proofs~\cite{trinh2024solving,hubert2025olympiad,georgiev2025mathematical}, they frequently generate opaque, uninterpretable constructions that provide little conceptual insight into \emph{why} a result holds~\cite{davies2021advancing,tao2026mathematics}.
This has triggered a widely discussed epistemic tension, since without deliberate human-in-the-loop interfaces, mathematical discovery risks drifting from a creative, understandable process toward opaque, automated search.
Finding the optimal way for domain experts to harness this potential remains a complex challenge that demands rigorous interface design~\cite{morris2025hci, amershi2019guidelines}.

Scientific discovery with AI operates across two interconnected loops: an autonomous \emph{inner search loop}, where the underlying agent iteratively generates, executes, and evaluates thousands of candidate solutions; and an \emph{outer scientific discovery loop}, where the domain expert defines hypotheses, configures constraints, and interprets findings.
While early conversational coding assistants focused on rapid, single-turn code generation for well-specified tasks, scientific discovery systems execute long-horizon inner loops over days or weeks on problems that are inherently open-ended and not fully knowable at the outset.

Despite the rapid progression of autonomous inner loops, current systems designed for automated scientific discovery (\eg{} Kosmos~\cite{mitchener2025kosmos}, AI co-scientist~\cite{gottweis2025towards}, and AlphaEvolve~\cite{novikov2025alphaevolve}) often treat their underlying models as \emph{opaque boxes} where the user enters a query and awaits a result, with human intervention relegated to the high-latency space between batch runs.
This creates a critical bottleneck in the outer scientific loop: first, users receive little interactive support to frame and constrain their objectives in forms the search process can productively explore; and second, users are expected to parse and validate large amounts of returned data, generated at significant cost, before determining their next action.

To address this bottleneck, we designed and deployed an interactive interface for AlphaEvolve~\cite{novikov2025alphaevolve} aimed specifically at tightening the outer scientific loop.
We developed this interface alongside internal subject matter experts (SMEs) over a period of nine months and evaluated it with \numplaytesters{} external mathematicians over a period of \playtestduration{}.
During the \playtestname{}, \playtestername{}s used the system to tackle novel research problems in combinatorics, geometry, and probability.
The mathematicians who had access to AlphaEvolve created more than \numexperiments{} distinct experiments, successfully identified novel solutions to a range of problems in their specialized subject areas, and have since published multiple papers on their discoveries (\eg{} \cite{ellenberg2026bruhat, berczi2026evolving1, berczi2026evolving2}).

Based on our observations, we propose \emph{intentmaking} as a mode of operation distinct both from traditional static specification alignment and straightforward code completion.
When working with coding agents, the user's intent is largely known, and the challenge is in communicating it to the system.
In contrast, open-ended scientific discovery treats intent formulation as an experimental process: the scientist iteratively builds intuition for the problem space, constructs a mental model of the behavior of their experimental apparatus (the AI), and eliminates systematic errors in pursuit of their results.

We define \emph{intentmaking} as this iterative process of discovering, framing, and refining one's problem specification by interacting with an AI system.
We see this as a natural counterpart to sensemaking, characterizing the outer scientific loop as an ongoing cycle where users alternate between intentmaking (defining, constraining, and updating their experiment) and sensemaking (interpreting and diagnosing results)~\cite{russell2024sensemaking}.
Together, this cycle repeats continuously, tightening the outer loop between mathematical hypothesis and empirical validation.
Rather than treating agentic models as autonomous black-box oracles that replace human cognition, this workflow reframes AI as an interactive scientific instrument that domain experts actively steer, calibrate, and interrogate to expand their own research capabilities.

\section{Related Work}

This work bridges two intersecting domains: the methodological evolution of \emph{AI support for scientific discovery}, and the HCI domain of \emph{defining intent and analyzing results for agentic systems}.

\subsection{AI Support for Scientific Discovery}
The integration of computer aid into scientific research has a long history of enabling and accelerating discovery~\cite{appel1977every,hales2006formulation}.
The intellectual lineage of the ``AI Scientist'' paradigm begins with early expert systems, such as the DENDRAL project in the 1960s, which utilized task-specific heuristic programming to infer molecular structures~\cite{lindsay1993dendral}, and the BACON systems of the 1980s, which relied on data-driven search heuristics to rediscover quantitative empirical laws~\cite{langley1981bacon}.
While foundational, these early symbolic systems were often limited by the combinatorial explosion of modern scientific data.

Today, computation-supported research is ubiquitous across scientific domains~\cite{wang2023scientific} and modern AI for science has brought specialized methodologies that drive significant breakthroughs.
Reinforcement learning has formalized proof search in automated theorem proving~\cite{hubert2025olympiad,trinh2024solving}.
Bayesian optimization and active learning are routinely deployed in self-driving laboratories to navigate high-dimensional experimental parameters with minimal physical trials~\cite{kusne2020fly}.
Additionally, symbolic regression frameworks utilize reinforcement learning and Bayesian methods to uncover human-interpretable mathematical expressions from noisy observational data, bypassing the limitations of traditional genetic programming \cite{petersen2019deep,guimera2020bayesian,boussif2025bayesian}.

AlphaEvolve~\cite{novikov2025alphaevolve} is part of an emerging family of systems that leverage test-time compute~\cite{gottweis2025towards} to spend additional inference resources exploring solution spaces.
The underlying mechanics of AlphaEvolve build upon the rich intersection of Evolutionary Computation (EC) and LLMs, where language models act as intelligent mutation operators to guide evolutionary search toward promising regions of a solution space~\cite{liu2023evolutionary,romera2024mathematical}.

Related systems include Co-scientist~\cite{gottweis2025towards}, which is designed to formulate novel research hypotheses based on a user prompt, and Kosmos~\cite{mitchener2025kosmos}, which attempts to automate data-driven scientific discovery.
Concurrently, systems like ``The AI Scientist''~\cite{lu2024ai} and various Agent Laboratory frameworks~\cite{schmidgall2025agent} attempt to close the entire research loop autonomously, from literature synthesis to manuscript generation.

However, a critical limitation of pursuing complete inner-loop autonomy is that open-ended scientific frontiers are rarely solvable in a single turn.
Without human steering, autonomous agents struggle with sparse rewards and combinatorial exploration of irrelevant subspaces.
In practice, current systems force users into coarse-grained interaction: submitting a high-level question, waiting through an opaque computational run, and sifting through vast dumps of returned data.
In contrast, the AlphaEvolve user interface is designed around a fine-grained, collaborative loop in which domain experts iteratively adapt problem statements and evaluation criteria in response to partial discoveries, operating the system more like an interactive scientific instrument (akin to an electron microscope) than an opaque autonomous assistant.

\subsection{Intentmaking and Sensemaking}
Within the field of HCI, \emph{human-in-the-loop} is a well-explored concept~\cite{morris2025hci,mosqueira2023human} with implications both in the training, steering, and explainability of AI.
Our work can be considered a practical application of human-in-the-loop concepts in the field of scientific discovery, where agentic systems operate as a new UI paradigm~\cite{morris2025hci}.

The HCI community has investigated the analysis process of large or complex data under the term sensemaking~\cite{russell2024sensemaking}.
Classic sensemaking models describe iterative loops of information foraging, schema building, and hypothesis testing.
In complex AI interactions, linear conversational interfaces are often insufficient; systems like Sensecape~\cite{suh2023sensecape} have demonstrated that providing multilevel abstractions and visual hierarchies is crucial to help users manage cognitive overload when processing extensive LLM outputs.
However, while traditional sensemaking focuses on interpreting external data or static model responses, agentic discovery requires a reciprocal steering loop: domain experts must actively constrain and re-specify the computational process in light of new insights.

In traditional HCI and design, the process of tackling ill-defined problems is often understood through Schön's concept of reflection-in-action~\cite{schon2017reflective}, where a practitioner engages in a continuous~\emph{conversation with the situation} to simultaneously understand, solve, and redefine a problem.
We propose that intentmaking represents the necessary evolution of this reflective practice when applied to agentic AI.
While traditional reflection-in-action involves manipulating static or predictable materials, intentmaking captures the unique cognitive friction of collaborating with a non-deterministic, opaque, and goal-oriented system.
Users are not simply refining a static specification; they must continuously co-evolve their problem definition alongside their growing mental model of the AI's behaviors, capabilities, and tendencies to reward-hack.
Thus, based on our observations when designing a UI for AlphaEvolve, we highlight intentmaking as an integral but not yet well-studied aspect of working with agentic systems.

Our work is closely related to Interactive AI Alignment~\cite{subramonyam2024bridging,terry2023interactive,feng2024cocoa}, which highlights alignment work in the specification, process, and evaluation phases of working with an agentic AI system.
In this paper, we add to existing work around AI design guidelines~\cite{amershi2019guidelines} and argue that users typically do not fully know how to describe their objective before getting started, and instead rely on experimentation to help them define it.

The proposed intentmaking process, which maps Horvitz's principles of mixed-initiative user interfaces~\cite{horvitz1999principles} to AI-assisted discovery, is especially important when bringing AI assistants to domain experts who might have a less clear mental model of the AI's capabilities compared to AI experts~\cite{zamfirescu2023johnny,jiang2022promptmaker,zamfirescu2023herding,joshi2025coprompter,yang2020re}.
We observe a bidirectionality in the process of hypothesis, experimentation, and refinement enabled by our system, similar to what is described in Shen~\ea{}'s position paper on bidirectional alignment between humans and AI systems~\cite{shenposition}.
However, Shen's work is more focused on value alignment~\cite{leike2017ai,irving2018ai,christian2020alignment} than refining user intent.
\\
\\
In summary, our work is at the intersection of AI for scientific discovery~\cite{wang2023scientific} and research on HCI paradigms when interacting with AI systems~\cite{morris2025hci}.
We observe that when using agentic AI systems for open-ended discovery, formulating intent is an iterative process of intentmaking and sensemaking, requiring the user to continuously build and update their mental model of the system, steer its search landscape, and refine their problem specification.

\section{Interacting with AlphaEvolve}

AlphaEvolve \cite{novikov2025alphaevolve} is an evolutionary coding agent that orchestrates a pipeline of LLMs to iteratively optimize a piece of Python code.
In its simplest form, user interaction with AlphaEvolve requires the user to express their goal with three artifacts:

\begin{enumerate}
    \item \textbf{Problem Statement}: The problem statement describes the goal of the experiment and scientific problem in natural language.
    It is used as an augmentation to the prompt given to AlphaEvolve and helps the model understand the user's intent.
    \item \textbf{Initial Program}: Typically a naive, non-optimized solution to the problem.
    This serves as the starting point for evolution from where the first candidate is created by asking the model to produce a code diff.
    \item \textbf{Evaluation Function}: A function called for every generated candidate that returns scores for a user-defined set of metrics.
    This function cannot be modified by the AlphaEvolve system.
\end{enumerate}

Based on this experimental specification, AlphaEvolve applies evolutionary optimization techniques to iteratively propose and evaluate improved solutions.
The results of this process are returned to the user through a running \emph{best score} and the candidate responsible for that score for each metric.
Using this approach, AlphaEvolve has achieved a significant number of state-of-the-art results \cite{georgiev2025mathematical}.
However, the practical application of this process is limited by a number of complexities.
During observational studies with internal SMEs, we identified the following key themes:

\begin{enumerate}
    \item \textbf{Underspecified Constraints}:
    While users typically started with a clear mathematical objective (\eg{} \plainquote{maximize the number of unit spheres that can be packed around another unit sphere}), there remained many unknowns around how to approach and define this problem.
    Analogously, in classical scientific workflows, many details of the proposed experiment often only surface as experimentation takes place, and crystallize in the light of initial results.
    \item \textbf{Incompatible Problem Statements}:
    The user-conceived problem statement (\eg{} \plainquote{please find a new lower bound for kissing numbers in 24 dimensions}) often does not align with the patterns that enable AlphaEvolve to discover an optimal solution.
    Viewed through an HCI lens, this represents a Gulf of Envisioning~\cite{subramonyam2024bridging}.
    Users struggle to anticipate the literal or pathological ways an agentic system might interpret a mathematically sound objective.
    Consequently, SMEs expended significant effort over many months engaging in intentmaking --- iteratively bridging this gulf by discovering the hidden constraints and reward landscapes that best guided AlphaEvolve toward valid solutions.
    \item \textbf{Meaningless Progress}:
    The LLM-driven process that generates program candidates can sometimes find solutions that improve the score in non-functional ways, \eg{} by returning an unexpected type in order to trick the evaluation function into providing a high score.
    Similarly, the LLM-driven process can get caught in non-productive loops, perhaps incrementally improving on the score by virtue of random chance without making real progress.
\end{enumerate}

Based on these insights, we developed an interactive interface for creating and monitoring AlphaEvolve experiments.
In the following, we will describe this system through the lenses of the intentmaking and sensemaking processes, which together form an iterative experiment specification and monitoring loop.

\begin{figure}[ht]
  \centering
  \includegraphics[width=0.6\linewidth]{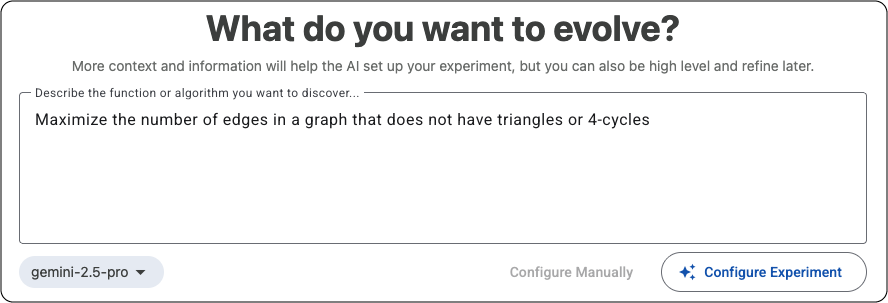}
  \caption{Initial prompt for the user to enter a description of the problem they want to solve.
  The \setupagent{} will use this description alongside a system prompt to scaffold a new experiment.}
  \Description{Single-input text box interface prompting the user to enter a high-level description of the mathematical problem they wish to explore with AlphaEvolve.}
  \label{fig:problem_description}
\end{figure}

\subsection{Intentmaking}

To understand this workflow, we formalize the concept of intentmaking.
Drawing from Schön's reflection-in-action~\cite{schon2017reflective}, where problem-solving and problem-setting occur simultaneously, intentmaking applies this reflective practice to open-ended scientific discovery.
In this paradigm, users do not simply execute a pre-defined plan; instead, they engage in an iterative, experimental loop of observing the AI's behavior, discovering unforeseen facets of their own experiment, and adjusting their mental model to align the AI's actions with their true, evolving goals.
In the context of AlphaEvolve, the intentmaking process applies both to experiment definition and the creation of variants of earlier experiments.

\begin{figure*}[ht]
  \centering
  \includegraphics[width=\textwidth]{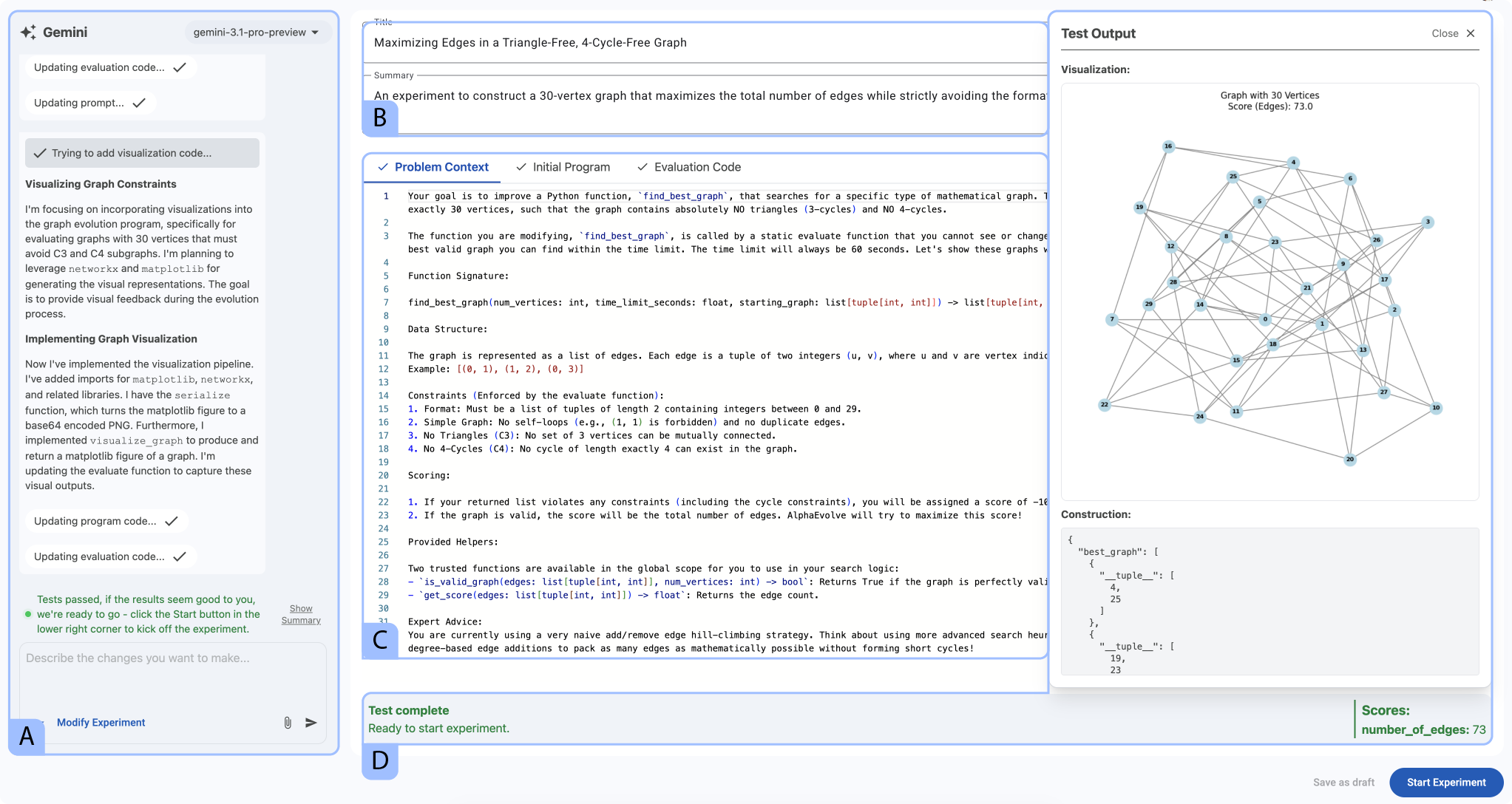}
  \caption{Experiment configuration assistant. The user describes their problem and may request changes to the experiment setup in a chat interface (A). An automatically generated title and description can be changed by the user (B). The experiment setup with its full configuration is generated and can be edited by the user (C). Before starting an experiment, users do a test run that assures them the experiment is set up as intended (D).}
  \Description{Split-pane experiment configuration assistant interface. The left pane shows an interactive conversational agent, while the right pane displays structured tabs for problem context, initial program code, and evaluation functions.}
  \label{fig:assistant}
\end{figure*}

AlphaEvolve experiments are defined in the form of a natural language problem definition, an initial program, and an evaluation function (both defined in Python).
We designed a minimal initial interface with a single high-level prompt familiar from many other systems (see \autoref{fig:problem_description}), and an experiment definition process that resembles traditional agentic code editors (\autoref{fig:assistant}). 
This experiment definition UI consists mainly of a chat panel on the left (\autoref{fig:assistant}, A), and tabs for the three main artifacts needed to scaffold an AlphaEvolve experiment in the center (\autoref{fig:assistant}, C): the problem context, initial program, and evaluation function.

Several patterns have been developed to enable AlphaEvolve to perform optimally, but even experienced users do not have the same level of in-depth knowledge of the system as the core team.
To bridge this gap, we added this knowledge as a prompt to the \setupagent{} such that the system will guide the user to apply patterns that are best suited to AlphaEvolve to their own specific problem.
To illustrate the benefit of this, consider the challenge of finding lower bounds for kissing numbers in high dimensions, a fundamental problem in discrete geometry that asks for the maximum number of non-overlapping unit spheres that can touch a central unit sphere \cite{georgiev2025mathematical}.
Rather than asking AlphaEvolve to directly generate an optimal point cloud, SMEs discovered that the system performed better if the problem was framed as a time-bounded search which updated the initial points and the search function in tandem.
The experiment definition process captured this institutionalized knowledge and guided users to frame their own experiments in a form that AlphaEvolve was best equipped to tackle successfully.

Based on the initial specification provided by the user, and building on our learnings from SMEs, the coding agent proposes an initial experiment setup.
We use the agent to generate a title and description for the experiment in addition to the three main artifacts mentioned above.
This setup enables users who do not typically interact with code to set up their AlphaEvolve experiments from just an abstract problem specification.
Internal tests showed that neither the initial problem specification provided by the user, nor the generated experiment setup were always fully correct on the first try. In many cases, the provided problem specification lacked details which led to an underspecified experiment setup. To further help users advance in their intentmaking process, we designed this second stage of intentmaking with a focus on iteration and testing.

To make it easier for users to understand individual candidate solutions, we incentivize the \setupagent{} to generate not just evaluation metrics, but also visualization code.
Visualizations serve as a high-level boundary object between code and human understanding, allowing users to gain an intuitive understanding of a candidate's behavior without having to inspect the code directly.

Running a full evolutionary experiment in AlphaEvolve is computationally intensive, requiring thousands of LLM calls potentially over several days.
For researchers without access to massive compute clusters, launching an underspecified or flawed experiment is prohibitively expensive.
While some facets of a problem only emerge during long-horizon search, we designed a local test stage to eliminate common specification flaws during experiment definition.
In this stage, users execute their code and evaluation function within a sandbox, immediately inspecting the resulting metrics and visual constructions.
Coupled with a critique agent that scans the setup for known reward-hacking patterns, this allows users to iteratively debug and calibrate their intent before committing to a costly optimization process.

\subsection{Sensemaking}
AlphaEvolve can run for several days and generate many tens of thousands of programs which are often thousands of lines long. Clearly, presenting all these programs to a human user is of limited value. After discussing with internal SMEs and \playtestername{}s, we identified several patterns of usage which are covered in detail in \autoref{sec:results}, and we designed the experiment dashboard (\autoref{fig:dashboard}) to address those usage patterns.

At the highest level, users can see basic experiment metrics (generated programs, executed programs, failed programs), as shown in \autoref{fig:dashboard} (A).
AlphaEvolve experiments can fail for various operational reasons (\eg{} jobs may be preempted by other systems requiring resources, or LLM quota might be exhausted), and experiment metrics provide the first, simple sanity check that the experiment is still running.  

The current \emph{best score} that AlphaEvolve has found is often the key information users need to decide if an \emph{interesting result} has been achieved.
The \emph{Best Scores} shown in \autoref{fig:dashboard} (B) provides an indication of how scores have progressed over time.
Most graphs tend to have a similar shape, with rapid gains made early on, followed by gradual, asymptotic improvement.
The bar chart below the main graph provides an indication of \emph{how many} best programs have been identified as a function of time.
This provides users with an important indication of whether their experiment is still generating potential new solutions.

\begin{figure*}[ht]
  \centering
  \includegraphics[width=\textwidth]{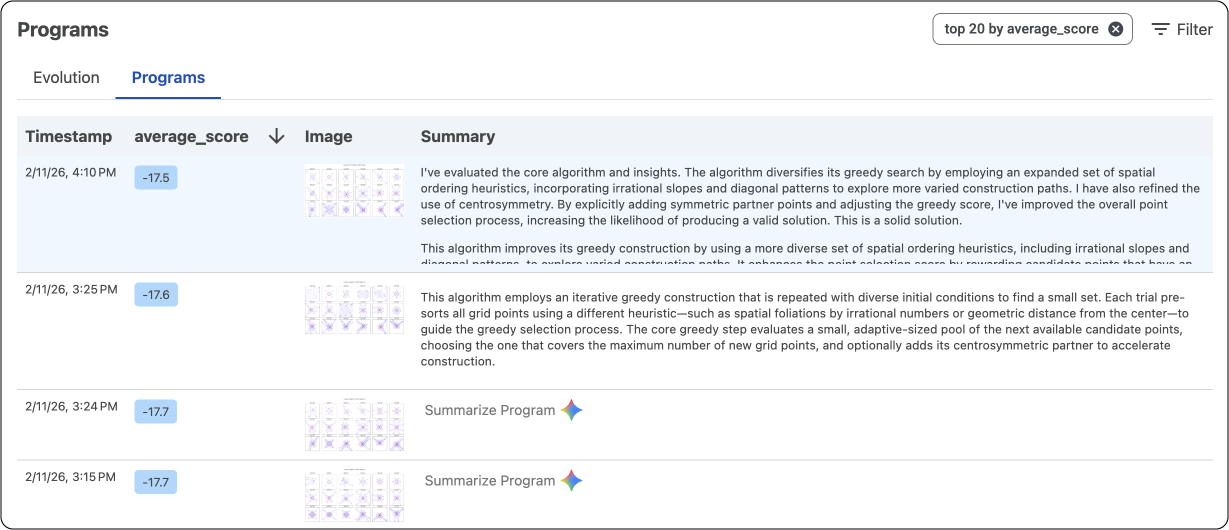}
  \caption{Programs Table showing candidate solutions, in this case sorted by the currently selected metric. The table shows images generated during evaluation and AI summaries can be generated to give the user insight into the candidate's approach.}
  \Description{Filterable and sortable program table displaying candidate solutions, their numerical evaluation scores, thumbnail visualizations of geometric constructions, and AI-generated summary explanations.}
  \label{fig:program-table}
\end{figure*}

The \emph{Programs Panel}, shown in \autoref{fig:dashboard} (C), provides up to three ways for users to reason about the generated programs \emph{as a whole}.
These are the \emph{AI Overview}, \emph{Evolution Tree}, and \emph{Programs Table} (\autoref{fig:program-table}).
We opted to experiment with multiple representations for this view since they all serve different use cases.
The AI Overview gives a high-level overview of what has been explored in the experiment, whereas the Evolution Tree allows users to investigate specific branches.
The Programs Table can be filtered and sorted, so it is best suited to explore the top programs for a given metric.

Selecting any program from the Best Scores or the Programs Panel causes it to be displayed in the \emph{Selected Program} panel.
This panel surfaces multimodal artifacts that allow domain experts to assess the state of the experiment without parsing program code, which is often not the best artifact with which to judge the success of an experiment.
These include the raw mathematical construction (\eg{} generated adjacency matrices or point clouds), visual representations of the construction, and AI-generated summaries of the candidate's algorithmic strategy (\cf{} \autoref{fig:program-table}).
By triangulating between the evolution tree, visual artifacts, and metric trajectories, domain experts can quickly make informed outer-loop steering decisions: determining whether a branch represents genuine mathematical progress, aborting runs that exploit metric loopholes, or extracting promising candidate strategies to fork into new, refined experiments.

\section{Design and Evaluation}\label{sec:setup}

The development and evaluation of the AlphaEvolve interface followed an iterative, human-centered design process comprising two distinct phases: a nine-month co-design and internal deployment phase with subject matter experts (SMEs), followed by a longitudinal evaluation with eleven external academic mathematicians over \playtestduration{}.

\subsection{Iterative Co-Design and Internal Deployment}
AlphaEvolve was initially deployed internally as a command-line interface (CLI).
While functional for AI researchers, the CLI created high operational friction: configuring experiments required manual script writing and environment setup, while monitoring multi-day runs required parsing dense text logs or querying databases directly.

Over a nine-month period, we worked closely with internal SMEs (including theoretical mathematicians and machine learning researchers) to transition from the CLI to an interactive web-based system.
This process followed multiple rapid design cycles:
First, we built an initial monitoring interface focused on high-level sensemaking metrics (\autoref{fig:dashboard}, A and B) to help users verify job health and track score progression.
Second, based on SME feedback regarding the difficulty of manual configuration, we developed the \setupagent{} and the pre-execution test sandbox (\autoref{fig:assistant}).
At the time of our external evaluation, the internal deployment supported approximately \numactives{} monthly active users across diverse scientific domains.

\subsection{External Expert Evaluation}
While internal deployment provided continuous feedback, internal users had deep prior familiarity with AlphaEvolve's architecture and direct access to the core engineering team.
To evaluate how domain experts independently engage with the system on open research problems, we conducted a longitudinal evaluation with external mathematicians.

\paragraph{Participants.}
We recruited \numplaytesters{} \playtestername{}s consisting of faculty and senior researchers in mathematics from leading universities across the US and Europe.
Their self-reported programming experience ranged from purely theoretical researchers who rarely write code to computational mathematicians proficient in Python.
All participants provided informed consent in accordance with our institution's ethical research standards.
All interview transcripts, interaction logs, and qualitative feedback were strictly anonymized.
Participants whose mathematical discoveries resulted in public manuscripts provided explicit consent for their publications to be cited as evaluation outcomes.

\paragraph{Study Procedure and Support.}
The study spanned \playtestduration{}.
Each \playtestername{} began with an individual 45-minute onboarding session where we introduced the interface and assisted them in formulating their first research question.
Following onboarding, participants had unrestricted access to the system to run experiments on problems of their choice.
To support realistic, asynchronous collaboration and capture in-the-moment feedback, we established a shared communication channel connecting all participants with the system designers and mathematicians from the core development team.
After approximately three to four weeks of active usage, we conducted 45--60-minute semi-structured follow-up interviews with participants to explore their workflow, sensemaking strategies, failure modes, and interaction friction.

\paragraph{Data Collection and Analysis.}
Our analysis triangulates across three data sources:
\begin{enumerate}
    \item \textbf{Interview Transcripts}: Audio recordings of onboarding and follow-up interviews were transcribed.
    \item \textbf{Asynchronous Interaction Logs}: We gathered all conversational threads from the shared discussion channel, capturing real-time problem-framing discussions, reward-hacking diagnoses, and feature requests.
    \item \textbf{System Telemetry and Artifacts}: We logged all \numexperiments{} experiment configurations, starter programs, evaluation functions, execution runtimes, and resulting candidate trees.
\end{enumerate}

Two researchers independently conducted open inductive coding on interview transcripts and communication logs, identifying recurring patterns in how mathematicians formulated objectives, inspected candidates, and navigated failures.
The researchers regularly compared codes, resolved discrepancies, and organized codes into higher-level themes centered on the intentmaking--sensemaking cycle, reward-hacking mitigation, and workflow bottlenecks.
Quantitative telemetry data was used to contextualize these qualitative themes.

\section{Results}\label{sec:results}

Over \playtestduration{} of usage, our \numplaytesters{} \playtestername{}s created over \numexperiments{} experiments 
focused on their own areas of expertise (combinatorics, geometry, and probability), and published multiple papers based on their findings (\eg{} \cite{ellenberg2026bruhat, berczi2026evolving1, berczi2026evolving2}).
In the following, we will present the qualitative results of this \playtestname{} obtained through user interviews and observations as described in \autoref{sec:setup}. We identified a number of usage patterns that we believe are generally applicable across AI scientist systems.

\subsection{Intentmaking}

A primary goal of our interface was to lower the barrier to entry, particularly for domain experts who do not write code daily. 
Our \playtestername{}s told us that, with the \setupagent{}, \plainquote{the barrier to getting started is so low, I just think of something and I think `well, let's try that'.}

For many \playtestername{}s, a lower barrier to entry meant that they could try out ideas they might not otherwise have had the confidence to invest time in.
One participant commented on the enabling nature of the tool: \plainquote{The fact that I don't have to write any of the code myself is very, very nice ... if I had to write this code myself, I would have never really gotten anywhere.}

Aligned with AI-powered coding agents, users did not expect to get it right on the first try; rather, they expected a conversational, iterative refinement process.
As one expert summarized: \plainquote{I think I take it to be a given that there's going to be an iterative process where I set it up and then I'm like oops no I want to change it and I don't want to start over.}
During experiment definition, we observed that users heavily relied on the low-cost test stage to iterate rapidly with visual feedback.
Because setting up an experiment only required a high-level prompt, one user noted they \plainquote{deliberately didn't spend too long trying to think about what I wanted it to do... I thought it would be easier to see what it guessed and then try and correct it.}
Users treated the initial AI-generated specification as a disposable starting point, running fast, local tests to evaluate the generated code and visual feedback before safely committing to a full experiment.

\begin{figure*}[ht]
    \centering
    \includegraphics[width=\textwidth]{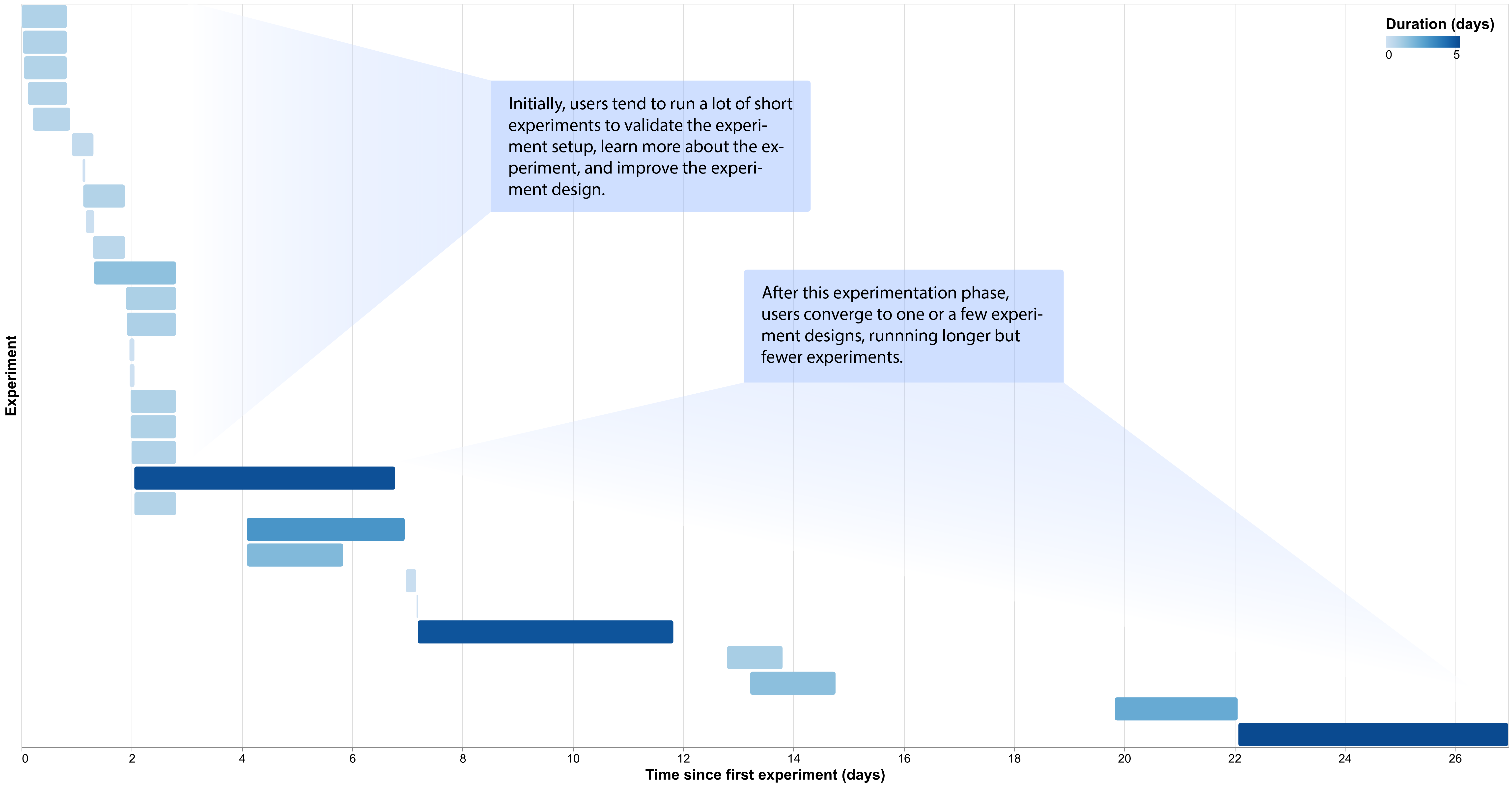}
    \caption{Experiment start and stop times for a representative \playtestername{}, showing the process of starting with short-lived experiments and gradually refining the experiment setup before running multi-day experiments. The \playtestername{} had several such runs going in parallel, working on different problems, but this graph only shows experiments for the same matrix connectivity problem for illustrative purposes.}
    \Description{Gantt chart displaying experiment start and stop times over several weeks, illustrating how users iteratively launch short exploratory runs before committing to longer multi-day experiments.}
    \label{fig:experiment-timings}
\end{figure*}

Furthermore, although some users still painstakingly \plainquote{copy pasted a very detailed description of the problem,} most gave a very lightweight initial problem statement in the expectation of refining it during the definition stage.
Despite this rich experiment definition stage, users rarely ran just a single experiment for a given problem (\cf{} \autoref{fig:experiment-timings}).
This experimental interaction is a key observation that motivates our concept of an intentmaking--sensemaking loop.  

The following dialogue between one of the \playtestername{}s and a member of the team gives a good flavor of this process, in which both the characteristics of the problem and the behaviors of AlphaEvolve have to be considered together in order to refine the experiment definition.

\begin{dialogue}
    \speak{Team} I think the reason it's not working well is that the reward is too sparse. It's easy to do $n$ (or $n-1$ with your hint) but after that it just keeps trying new ideas without knowing whether or not it's going in the right direction.
    \speak{User} Is there a way to add partial scores, so that a score of 5.1 means 6 hyperplanes, but it's close to 5?
    \speak{Team} Maybe if 5 hyperplanes cover almost all points, that's better than having 6 hyperplanes covering $1/6$ each?
    \speak{User} A natural one is slicing most of the edges, but it's the kind of problem where the 99\% case is very different from the 100\% case.
    \speak{Team} If we set the penalty for missing edges pretty high, then it will have flexibility to move around while still prioritizing 100\% constructions.
\end{dialogue}

In this conversation, the team discusses how to compensate for a sparse reward function.
This illustrates that intentmaking in scientific discovery extends beyond prompt wording into \emph{reward landscape shaping}, modifying scores, penalty terms, and secondary ``cross-check'' metrics (\eg{} \plainquote{I need the best score that has no violations}) to provide viable search gradients and rule out obviously pathological results.

\subsection{Sensemaking}

Immediately after experiment launch, and multiple times throughout the first day, users inspected the overall metrics to ensure that their experiment was running: \plainquote{A flat line on any of the key metrics indicates a problem}.
Thereafter, they tended to check in daily, or sometimes wait multiple days before next checking their experiment.
In this phase, users were looking at the overall shape of the score chart, evolution tree, constructions, and visualizations to determine if their experiment was likely to be successful if allowed to run for longer: \plainquote{I want to see if my experiment is making `real' progress}.
We also found out that users often know what answer they are looking for.
For example, if \plainquote{-0.5 is the interesting region, I can see it's getting closer but I don't think it's ever going to actually get there}. 
In such cases, helping users understand an experiment's trajectory was highly valued.

While programs are typically the main deliverable of AlphaEvolve, diagnosing the progress of the experiment by reviewing a 1,000+-line program is not efficient.
Users tended to rely on a range of supplementary information to determine if their experiment was progressing.
In many cases the score chart itself provided the most immediate clues (\eg{} whether increments were within the bounds of random chance and rounding error or whether larger strategic improvements were still being made); beyond the score chart, the constructions (data generated by the programs) and visualizations of the constructions were the most widely used artifacts.
Finally, we provided AI Summaries of the programs (\autoref{fig:program-table}, right). While these were just simple, single-shot AI-generated summaries and provided only vague overviews of the strategies used, they provided a valuable signal that helped users decide whether to investigate programs more closely. 

Users often had to check the robustness of candidate solutions, and \plainquote{let it run a bit longer because I wanted to see how other solutions evolved and whether simpler ones might emerge.}
This is important for three reasons: firstly, to rule out systematic errors and false-positives (\plainquote{I think there's a lot of sanity checking to be done on the things it gives me}); secondly, to determine whether a candidate is genuinely promising or merely serendipitous (\plainquote{I can't tell if this was just a lucky configuration or an interesting new strategy}); and thirdly, to extract the underlying mathematical idea if a novel strategy is suspected (\plainquote{to understand what this idea was [...] because this branch had lots of good ideas}) in order to generalize it and apply it to other problems.

\subsection{Reward Hacking}

Reward hacking is a form of specification misalignment in which the system maximizes the evaluation function without making useful progress towards the intended goal.
It is common in many reinforcement-learning and evolutionary systems, and a frequent failure pattern in AlphaEvolve experiments.
In the case of AlphaEvolve, reward hacks can be particularly ingenious because the system relies on LLM-powered code mutation.
For example, one user told us the system \plainquote{created a new class type list in which it updated the meaning of length.} 
Such reward hacks are notoriously hard to foresee, and \plainquote{it takes a lot of effort to close the loopholes and get it to search for the thing that you're actually looking for.}

The critique agent employed during experiment setup, as shown in \autoref{fig:eval-analysis}, provided a valuable safety net here.
Many common patterns that lead to reward hacks had already been identified by the AlphaEvolve team during development, and we captured these in a prompt.
As with experiment definition, where the prompt guided the users to find the most appropriate AlphaEvolve patterns to tackle their specific problem, the critique prompt was effective at identifying common reward hacks before the evolutionary process had started.

Despite the critique agent, it was still common for reward hacks to emerge as part of the evolutionary process.
As such, one user pointed out that often, when they \plainquote{thought it was doing well, it was cheating.}
Hence, attempting to treat these models as autonomous assistants proved to be a risky strategy; instead, participants succeeded through iterative calibration, discovering step by step \plainquote{how you actually specify the objective in a way that it's actually what you want.}
Users frequently launched an exploratory experiment and monitored its trajectory before restarting a modified version (\cf{} \autoref{fig:experiment-timings}).

\subsection{Workflow Friction: Versioning and Boundaries}

While the core interaction loop was successful, the \playtestname{} revealed friction points in experiment management and UI affordances.
Because intentmaking requires heavy iteration, users quickly generated dozens of slightly modified experiments, leading to cognitive overload.
\plainquote{Telling my experiments apart is really difficult and I have to click into the experiment to remind myself what was different,} one mathematician reported.
Users expressed a strong need for better version control and visual provenance across experiments.
One requested \plainquote{a better view like an overview a tree of the modifications ... and a better and more efficient way of restoring previous versions.}

Additionally, the conceptual boundaries of the system were not always clear to first-time users.
A \playtestername{} admitted they \plainquote{actually didn't know that we could edit manually... until yesterday} and weren't \plainquote{really sure they liked the separation between initial program and evaluation code.}
Furthermore, the agent's eagerness to rewrite code sometimes caused destructive edits, with another noting that asking it to update one function sometimes \plainquote{destroys the rest of the evaluation code.}
This highlighted a critical interaction requirement: supporting non-destructive conceptual dialogue, allowing domain experts to brainstorm hypotheses and explore design trade-offs with the assistant without triggering automatic, destructive code overwrites.

\section{Discussion}

A classic human-centered goal for AI is to augment human intelligence rather than replace it, but the patterns that enable the practical implementation of this goal remain subject to open research.
In many ways, AlphaEvolve lends itself well to this because computer programs are, by their nature, easier to automatically validate.
This makes it simpler to filter the vast amounts of generated data into human-sized chunks and to extract structured data that will allow users to derive meaning more easily.
More general AI scientist tools that deal in human-language generation (\eg{} hypothesis generation) face a more challenging task, although the overall patterns explored here remain relevant.

Based on the findings from the user study and our discussions with expert mathematicians, we summarize below a set of design considerations for future user interfaces for AI-driven scientific discovery systems.

\subsection{Elevating Intentmaking and Iteration}

Getting the most out of a complex system like AlphaEvolve requires the user to develop a very sophisticated mental model of the system~\cite{pairguidementalmodels}, which we observed users building primarily through trial and error.

\paragraph{Lowering the barrier to entry.}
For scientists, the simplicity of being able to try out new ideas is not just a question of convenience; it is a vitally important contribution that any AI scientist system should prioritize.
In the course of daily work, scientists have many ideas, all with varying levels of potential impact.
The practicalities of academic and industrial time and budget constraints mean that scientists continually have to make choices about where to invest their time based on little more than their expert intuition, which inevitably leads to many potentially important ideas remaining \emph{on the shelf}.

A primary goal of an AI scientist system should be to enable the rapid, low-cost validation of ideas, thus enabling scientists to try out ideas they would otherwise have had to leave untested.
We believe this is an important shift in perspective, from \emph{asking questions} to \emph{trying out ideas}.
It is possible that the single most effective way that AI will accelerate science in the near term is not by generating novel ideas that humans could not have conceived themselves, but by making \plainquote{the barrier to getting started so low, I just think [...] 'well, let's try that'.}
However, there is an important caveat: the barrier to entry must consider not just the cost of making the initial query, but also the cost of validating the answers and then iteratively refining the question.
As such, we see the primary user interaction with AI-assisted scientific discovery systems as an intentmaking loop where the problem specification (intent) evolves alongside the candidate solution.

\paragraph{Defining and Elevating a Boundary Object.}
To give users a high-level entry point to defining and understanding the experiment setup, it helps to define a naturally understandable boundary object.
In the case of AlphaEvolve, this could be the problem context that is used to instruct new program generations, but the form of this boundary object generally depends on the type of experiment and user at hand.
Users must intuitively understand this boundary object, and modifying it should facilitate the definition of and modifications to an experiment.

\paragraph{Generalization to Modern Agentic Software Engineering.}
While our study investigates mathematicians, the dynamics of intentmaking and sensemaking also increasingly characterize modern software engineering with autonomous coding agents.
Recent empirical studies show that programming with AI shifts developer activity from purely generative coding to exploratory modes~\cite{barke2023grounded} and high-cost verification of model-generated code~\cite{mozannar2024reading}.
As coding agents evolve from single-turn autocompletion to multi-turn, long-horizon tasks, developers rarely operate with fully static, upfront specifications.
Instead, developers engage in continuous intentmaking and iteratively define test harnesses, constrain agent search spaces, and refine prompts.
Simultaneously, developer effort shifts from syntax generation to sensemaking: diagnosing multi-file diffs, understanding agent reasoning trajectories, and verifying design decisions.

\subsection{Enhancing Sensemaking and Diagnosis}

Effective use of complex agentic systems like AlphaEvolve requires the user to have a deep, intuitive understanding of how that system works, born of direct experience, \ie{} a \emph{tacit knowledge} rather than a rules-based one.
This is particularly true with the kind of LLM-powered agentic systems that are currently popular in AI scientist tools.

\paragraph{Prioritize the Experimental Loop.}
AI scientist systems should be architected around an iterative intentmaking--sensemaking loop that more closely resembles the traditional experimental method.
Crucially, we observed a nuanced behavior where user interaction with an AI scientist is not a single linear investigation, but two distinct forms of experimentation occurring in parallel: \emph{domain experimentation}, where users discover previously unknown characteristics of the problem they are investigating, and \emph{instrument experimentation}, where users understand and adapt to the behavioral dynamics, failure modes, and reward-hacking tendencies of the AI system itself.
In classical laboratory science, researchers divide their effort between investigating natural phenomena and calibrating, debugging, and ruling out systematic errors in their experimental apparatus.
Similarly, in AI-assisted discovery, adjusting constraints, tweaking penalty terms, and forking experiments are the primary mechanisms through which scientists calibrate the instrument to extract genuine discovery.
Discovery interfaces must therefore treat experiment restarts, branching, and metric recalibration as first-class, frictionless operations.
At a high level, the following quote from one of our \playtestername{}s acts as a useful north star for future design:
\plainquote{I feel like I could just go down a rabbit hole of playing with this thing all day... This is kind of like playing a computer game, right? We're sort of exploring the world.}

\paragraph{Elevate Program Constructions and Visualizations.}
Users primarily assess a solution by \emph{eyeballing} higher-level outputs such as constructions and visualizations.
We found that such a process is much more efficient than trying to understand the full solution for every AI-generated candidate.
This is especially important if the solution includes artifacts in which the user is not an expert, such as generated code for mathematicians.

\paragraph{Invest in Understanding and Debugging.}
Users need help to quickly grasp the strategy of a candidate solution within an experiment.
Automatic summaries and explanations of strategies can be provided across the solution space to aid in tracing ideas and understanding fundamental differences between candidate solutions.
\begin{figure}[ht]
    \centering
    \includegraphics[width=0.4\linewidth]{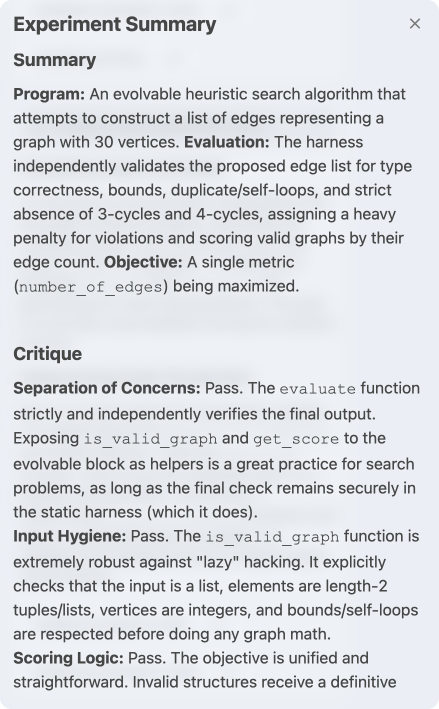}
    \caption{Automated diagnostic summary generated to validate the experiment design.
    This summary is generated when testing an experiment setup.
    It describes the setup in natural language and includes checks for design patterns based on expert feedback.}
    \Description{Automated diagnostic summary panel evaluating the user's experiment configuration, checking for common reward-hacking patterns, and providing natural-language suggestions for improvement.}
    \label{fig:eval-analysis}
\end{figure}

\paragraph{Provide Tools to Debug Metrics.}
Since reward hacking is a constant problem when working with AI assistants, interfaces must provide robust tools to diagnose issues with the defined metrics of success.
This includes support for multi-objective evaluation (\eg{} Pareto fronts) and synthesizing experiment progression insights to draw conclusions on potential metric flaws, supported by the aforementioned visual understanding tools.
One approach we explored was to capture the same kind of institutional knowledge that allowed problem statements to be mapped onto AlphaEvolve patterns in a \emph{cross-check} prompt that attempted to identify flaws in the experiment design at test time (\cf{} \autoref{fig:eval-analysis}).
Users found this helpful to identify flaws in their evaluation functions more quickly, thus tightening the experimental cycle, but more work is needed to ensure that the advice generated here can be mapped onto improved experiment performance.

\subsection{Fostering Robust Interaction}

Our findings suggest the interaction model should be robust to the non-deterministic nature of AI-assisted discovery, enabling users to maintain control and ask conceptual questions.

\paragraph{Robust Provenance and Versioning.}
Keeping track of many iterations of an experiment is a significant bookkeeping burden on scientists, and cataloging that knowledge in a separate system is likely to lead to duplicated effort and inaccuracy.
With substantial computational resources invested in AI-driven discovery, systems should invest equally in rich experiment tracking and provenance functionality.

\paragraph{Support Conceptual Inquiry.}
Allow users to ask conceptual questions of the AI assistant without aggressively or automatically modifying the experiment or the underlying code. This supports brainstorming and non-destructive refinement of intent.

\subsection{Navigating the Epistemic Shift in Mathematics}

The mathematical community is currently grappling with the broader implications of AI-assisted discovery~\cite{castelvecchi2023will,buzzard2023point,tao2026mathematics}. 
As automated and agentic systems achieve state-of-the-art results on open conjectures~\cite{romera2024mathematical,georgiev2025mathematical}, they introduce an acute epistemic dilemma: mathematics fundamentally values conceptual explanation, structural insight, and human understanding over raw computational verification~\cite{davies2021advancing,tao2026mathematics}.
In an emerging era of ``proof abundance'' without guaranteed human comprehension~\cite{tao2026mathematics}, an AI that outputs an opaque, 1,000-line program or an unmotivated coordinate matrix answers a query without advancing mathematical understanding.

Our observations demonstrate that the intentmaking--sensemaking loop provides a practical framework for resolving this tension. 
Through \emph{intentmaking}, mathematicians formalize, constrain, and stress-test their mathematical intuitions as search spaces and evaluation metrics.
Through \emph{sensemaking}, they do not merely check whether a score improved, but inspect visual constructions, trace evolutionary lineages, and extract structural patterns from candidate programs to formulate new human-understandable conjectures.
By shifting the interaction paradigm from an autonomous oracle to a steered scientific instrument, domain experts maintain intellectual agency and translate computational search into genuine mathematical understanding.

\section{Limitations}

We believe that many of our findings generalize to other AI scientist systems, but we acknowledge that our narrow focus on AlphaEvolve as a system, and mathematics as a domain, limits the generalizability of some of our recommendations.
Specifically, for problem domains where results are not as automatically verifiable or where iteration turnaround times differ significantly, we anticipate that further research is needed to identify better boundary objects and richer diagnostic tools that facilitate an effective intentmaking--sensemaking loop.

Within the time available, we explored a number of ways to support users to better frame their questions and understand the system's responses, but in many cases the solutions we built were simple first cuts based on single-shot AI prompts, and empirical evidence for their effectiveness is limited.
One area that would benefit from further work is in the diagnostic tools provided to the user.
There is an obvious synergy between the tendency of agentic systems to generate large amounts of data and the ability of LLMs to summarize data, but determining how to represent results in a way that is manageable for a user without risking the loss of potentially significant insights remains an open question.

Furthermore, we must acknowledge that this study was conducted within a highly resourced environment.
While our interface incorporates features like the test stage and the \setupagent{} specifically to minimize unnecessary compute cycles, the participants in our expert evaluation did not face strict financial or quota constraints when launching their experiments.
In broader academic and industry settings, compute access remains a significant bottleneck.
Future research must investigate how user behavior and sensemaking strategies change under strict resource constraints, and how UI affordances might better communicate estimated compute costs and budget management directly within the interaction loop.

\section{Conclusion}

In this paper, we presented the design and evaluation of an interactive interface for AlphaEvolve, an evolutionary coding agent for scientific discovery.
Co-designed over nine months with internal domain experts and evaluated in a \playtestduration{} study with \numplaytesters{} academic mathematicians, our system supported over \numexperiments{} experiments and enabled multiple published mathematical discoveries.
Our findings suggest that even with highly autonomous, long-horizon AI systems, deliberate interface design plays an essential role in making these tools effective for open-ended scientific research.

We proposed the \emph{intentmaking} and \emph{sensemaking} loop as a foundational framework for AI-assisted scientific discovery, tightening the collaborative outer loop by coupling interfaces that facilitate problem specification with visual diagnostic tools.
Ultimately, our work suggests that the most profound contribution of AI to science may not be autonomous idea generation, but drastically lowering the cost of \emph{trying out ideas}, and that conceptualizing AI scientist systems as interactive \emph{scientific instruments} may be a more productive paradigm than the traditional notion of an autonomous \emph{scientific assistant}.

\begin{acks}
We thank all internal and external collaborators for the extensive feedback they provided for this work.
We especially thank Gabi Cardoso, Anna Trostanetski, Victoria Johnston, Yosuke Ushigome, Richard Green, and our early external mathematical partners.
\end{acks}

\bibliographystyle{ACM-Reference-Format}
\bibliography{references}

\end{document}